\pdfoutput=1
\documentclass[preprint,12pt]{elsarticle}
\usepackage{graphicx}
\usepackage{amsmath,amssymb,amsfonts}
\usepackage{algorithmic}
\usepackage{algorithm}
\usepackage{textcomp}
\usepackage{lineno}
\usepackage{hyperref}
\hypersetup{hidelinks=true}

\journal{Robotics and Autonomous Systems}

\begin{document}

\begin{frontmatter}

\title{A Novel Path-Tracking Algorithm for Automated Tractor-Trailer Forward and Backward Maneuvers}

\author[utbm]{Alexandre Lombard\corref{cor1}}
\ead{alexandre.lombard@utbm.fr}
\author[utbm]{Florent Perronnet}
\author[utbm]{Th\'eo Rigal}
\author[utbm]{Nicolas Gaud}
\author[utbm]{Abdeljalil Abbas-Turki}

\cortext[cor1]{Corresponding author}

\affiliation[utbm]{organization={UTBM, CIAD},
            postcode={F-90010},
            city={Belfort cedex},
            country={France}}

\begin{abstract}
Fully autonomous tractor-trailer systems are increasingly deployed in logistics, agriculture, and industrial environments, where precise and robust path-tracking capabilities are essential. However, the articulation between the tractor and the trailer introduces additional nonlinearities and significantly complicates lateral and longitudinal control, particularly during reversing maneuvers. This paper introduces a novel path-tracking algorithm specifically designed for articulated vehicles with a single trailer. The proposed method combines a lateral control law applied at the trailer level with a short-horizon predictive adjustment of the tractor steering angle, ensuring stable convergence toward the desired path in both forward and backward motion. The approach is geometry-based and requires no per-vehicle calibration or training. Simulation studies in a high-fidelity physics simulator show steady-state or final lateral errors of $0.02$--$0.05$~m on straight-line, circular, and reverse docking trajectories, which is of the same order of magnitude as, or lower than, the errors reported in the literature. Real-world experiments on a small-scale tractor--trailer robot, with the whole controller running on a low-cost microcontroller, further confirm the feasibility of the approach, with lateral errors below 13~mm during repeated reverse docking maneuvers despite a pose detection latency of about 200~ms.
\end{abstract}

\begin{highlights}
\item Two-stage path-tracking controller for tractor-trailer systems
\item Trailer lateral control is converted into a target articulation angle
\item Forward correction maneuvers with hysteresis help recover the trailer angle
\item Validated in a high-fidelity simulator and on a small-scale real robot
\item The whole controller runs on a low-cost microcontroller at about 16 Hz
\end{highlights}


\end{frontmatter}

\section{Introduction}
\label{sec:introduction}

Autonomous tractor--trailer systems (TTS) are becoming increasingly relevant for a wide range of applications, including agricultural machinery, logistics platforms, automated yard operations, and industrial transport \cite{bechar2016agricultural}. Their ability to handle significant payloads while operating in constrained, cluttered, or dynamically evolving environments makes them particularly attractive for automation. However, these advantages also impose stringent requirements on motion control, as even small tracking errors may lead to unsafe configurations, collisions, or loss of maneuverability. Among the core control functionalities required for such systems, accurate and robust path-tracking plays a central role, directly impacting safety, productivity, and operational feasibility.

In contrast to passenger vehicles, articulated systems composed of a tractor and one or more trailers exhibit fundamentally different kinematic and dynamic properties. The presence of a hitch joint introduces additional degrees of freedom and strong nonlinear couplings between the tractor and the trailer \cite{brock2019dynamic}. In particular, lateral and heading errors applied at the tractor level propagate to the trailer with a delay and may amplify over time, especially at low speeds or during tight maneuvers. These effects are further exacerbated during reversing operations, where the system may become intrinsically unstable if not properly controlled. As a consequence, path-tracking for tractor--trailer systems remains a challenging control problem, even under simplifying kinematic assumptions.

Classical path-tracking controllers initially developed for non-articulated vehicles, such as Pure Pursuit \cite{coulter1992implementation} or the Stanley controller \cite{thrun2006stanley}, are widely used due to their simplicity and computational efficiency. However, when directly applied to articulated vehicles, these methods often fail to achieve satisfactory tracking performance or stability \cite{yang2025review}. Their primary limitation lies in the fact that they regulate the tractor motion only, without explicitly accounting for the trailer dynamics or the hitch angle evolution. As a result, they may produce steering commands that are locally optimal for the tractor but suboptimal or even detrimental for the trailer trajectory.

To address these limitations, several approaches have been proposed in the literature, ranging from geometric extensions of classical controllers to optimization-based and model predictive control (MPC) strategies \cite{kayacan2015learning,kayacan2018experimental}. While MPC-based methods can explicitly handle articulation constraints and multi-state coupling, they often come at the cost of increased computational complexity and sensitivity to model inaccuracies, which may limit their applicability in real-time embedded systems or low-speed industrial platforms.

In this paper, we propose a novel path-tracking algorithm for tractor--trailer systems that aims to bridge the gap between simplicity and predictive accuracy. The proposed method relies on a two-stage control strategy. First, an optimal lateral command is computed at the trailer level, directly regulating the trajectory of the most constrained and safety-critical part of the system. Then, a short-horizon predictive simulation is used to infer the corresponding steering command for the tractor that best realizes the desired trailer motion while respecting the system kinematics. This decomposition allows the controller to explicitly account for the delayed influence of tractor steering on the trailer trajectory, without resorting to full-scale nonlinear optimization.

The main contributions of this work are threefold:
\begin{itemize}
    \item a two-stage path-tracking controller for tractor--trailer systems, in which the output of a lateral control law applied at the trailer level is converted into a target articulation angle, which is then realized by a short-horizon predictive computation of the tractor steering angle; any lateral controller can be used in the first stage, and the method only requires the vehicle geometry, without per-vehicle calibration or training;
    \item a driving-mode selection strategy for reversing maneuvers, which triggers a forward correction maneuver when the target articulation angle cannot be reached in backward motion, with two hysteresis thresholds that prevent oscillations between driving modes;
    \item a demonstration, through real-world experiments on a small-scale robot, that the complete controller can run on a low-cost microcontroller while tolerating the localization errors and latencies of a real system, complemented by simulations in a high-fidelity physics simulator.
\end{itemize}

The remainder of the paper is organized as follows. Section~\ref{sec:background} introduces the kinematic model of tractor--trailer systems used throughout the paper. Section~\ref{sec:related_works} reviews existing approaches and discusses their limitations. Section~\ref{sec:tractor_trailer_path_tracking} presents the proposed two-stage control algorithm. Section~\ref{sec:experimental_setup} details the experimental setup and the validation results obtained in simulation and on a small-scale robot. Finally, Section~\ref{sec:conclusion} summarizes the contributions and outlines directions for future work.

\section{Background}
\label{sec:background}

This section introduces the kinematic model of the TTS on which the proposed controller (Section~\ref{sec:tractor_trailer_path_tracking}) relies, together with the notations used throughout the paper.

\begin{figure}
    \centering
    \includegraphics[width=0.6\linewidth]{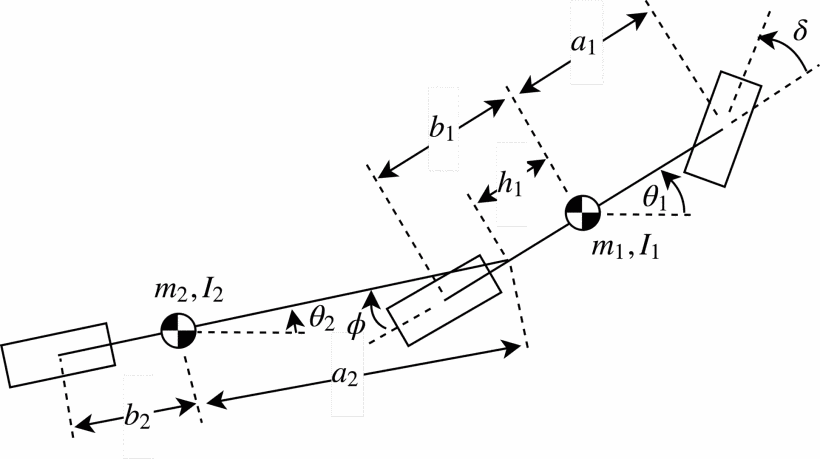}
    \caption{Extended kinematic bicycle model integrating the trailer, $\delta$ is the tractor steering angle and $\phi$ is the tractor-trailer angle}
    \label{fig:extended_bicycle}
\end{figure}

The extended bicycle model (Fig.~\ref{fig:extended_bicycle}) is a widely used representation to approximate the kinematic and dynamic behavior of articulated vehicles \cite{brock2019dynamic}. In its kinematic form, it captures the evolution of the tractor position, its orientation, and the articulation angle between the tractor and the trailer. The trailer motion is governed by the tractor orientation and the articulation dynamics, which are themselves affected by the steering angle of the tractor.

Let $\phi$ denote the articulation angle between the tractor and the trailer, $\delta$ the steering angle of the tractor, $L_1$ the tractor wheelbase, $L_2$ the distance between the trailer axle and the hitch point, and $v$ the longitudinal velocity at the hitch point (with $v>0$ in forward motion and $v<0$ in backward motion). The curvature of the tractor is given by $\kappa = \tan(\delta)/L_1$.
Under the no-slip assumption, the articulation angle dynamics can be written as
\begin{equation}
\dot{\phi} = v\left(\kappa - \frac{1}{L_2}\sin\phi\right).
\label{eq:articulation_dynamics}
\end{equation}

A limitation of this model is that it neglects tire slip and inertial effects, as the wheels are assumed to roll without slipping. The proposed controller therefore relies only on the geometry and kinematics of the TTS, and does not explicitly model its dynamics. This choice is consistent with the targeted applications, such as docking, yard operations, or agricultural maneuvers, which are performed at low speed and with moderate articulation angles: under these conditions, lateral tire forces remain small and kinematic models accurately describe the motion of the vehicle \cite{polack2017kinematic}. The remaining dynamic effects are handled as disturbances compensated by the feedback loop, and their influence is evaluated in Section~\ref{sec:experimental_setup}, both in a simulator that models tire forces and on a physical robot.

Classical path-tracking controllers designed for non-articulated vehicles, such as the Stanley controller \cite{thrun2006stanley} or its extension in \cite{lombard2020curvature}, assume a rigid vehicle geometry and do not account for the articulation dynamics described by \eqref{eq:articulation_dynamics}. The next section reviews the approaches that have been proposed to address this limitation for articulated systems.

\section{Related Work}
\label{sec:related_works}

Path-tracking for tractor-trailer systems has been extensively studied, and several recent surveys provide comprehensive overviews of the available control techniques \cite{yang2025review,song2026recent}. The proposed approaches can be broadly grouped into four families: geometric controllers, optimization-based controllers, learning-based controllers, and hybrid strategies.

\textit{Geometric and kinematic controllers.} For non-articulated vehicles, geometric controllers such as Pure Pursuit \cite{coulter1992implementation} and the Stanley controller \cite{thrun2006stanley} remain popular because of their simplicity and low computational cost, and comparative studies confirm that they offer a favorable accuracy-to-complexity trade-off in many situations \cite{artunedo2024lateral}. Several refinements improve their robustness to curvature variations and measurement noise \cite{lombard2020curvature}, or exploit the choice of the control point to enhance trajectory tracking \cite{lombard2026path}. For articulated systems, dedicated kinematic control laws explicitly account for the trailer motion, for instance by minimizing the trailer off-track along forward-followed paths \cite{michalek2015motion}. Sliding-mode formulations have also been applied to reverse path-tracking to cope with the unstable backward kinematics \cite{bin2023sliding}. These controllers are computationally light but tend to degrade under fast curvature changes, trailer swing, or actuator saturation.

\textit{Optimization-based controllers.} Model predictive control (MPC) is the most widely used framework for articulated vehicles, as it can explicitly incorporate articulation constraints, actuator limits, and trailer dynamics within a single optimization \cite{falcone2007predictive,kayacan2015learning}. Adaptive and speed-aware variants further improve tracking on curved or high-curvature roads \cite{kebbati2021optimized,Guan2024-ot}, and hybrid schemes combine MPC with geometric controllers such as Stanley to balance accuracy and robustness \cite{al2024new}. While MPC delivers excellent accuracy, its computational cost and sensitivity to model mismatch can hinder deployment on low-cost embedded platforms, which has motivated computationally lighter formulations for articulated vehicles, such as linearization-based MPC schemes \cite{kayacan2018experimental}.

\textit{Learning-based controllers.} More recently, reinforcement and deep-learning approaches have been investigated for lateral and path-following control \cite{kiran2022deep,li2019reinforcement,liu2024reinforcement}. They can capture complex, hard-to-model dynamics and adapt to varying conditions, but they typically require large amounts of training data, are sensitive to the sim-to-real gap, and provide limited stability or safety guarantees, which restricts their use on safety-critical articulated platforms.

\textit{Adaptive and hybrid strategies.} Hybrid strategies combining kinematic models with local optimization or adaptive laws have shown promising results, for instance in autonomous semi-trailer docking \cite{manav2021adaptive}, but many remain sensitive to model uncertainties, tire slip, or actuator saturation.

Overall, existing approaches face a recurring trade-off: lightweight geometric controllers struggle with the delayed and potentially unstable trailer dynamics, especially when reversing, whereas optimization- and learning-based methods achieve high accuracy at the cost of computational complexity, calibration effort, or limited guarantees. This motivates the development of a lightweight, geometry-based, yet predictive strategy that explicitly handles the trailer dynamics in both forward and backward motion, as proposed in this work.

\section{Tractor-Trailer Path-Tracking}
\label{sec:tractor_trailer_path_tracking}

\subsection{Setting the steering angle according to the trailer}

The proposed algorithm leverages the fact that the lateral trajectory of the trailer is primarily governed by the articulation angle between the tractor and trailer. The controller proceeds as follows.

First, a lateral control law $f_{lat}$ is applied at the trailer level. Using a classical lateral error and heading error formulation, we compute an expected articulation angle that would cause the trailer to converge to the desired path (this can be done with any lateral controller applied on the trailer). This step effectively transforms the trailer path-tracking objective into a target articulation angle.

Second, given this expected articulation angle, we search for a steering angle at the tractor level that reduces the difference between the current and desired articulation. To do so, we simulate the tractor-trailer kinematic model over a short time horizon, for a range of candidate steering angles. For each candidate, we compute how the articulation evolves and evaluate the distance to the target articulation.

Third, this process is repeated iteratively over the prediction horizon. At each step of the simulation, we refine the candidate steering angle based on the predicted evolution of the articulation angle.

Finally, the controller outputs the average steering angle computed across all simulated steps. This averaged value ensures a smooth and stable command for the tractor while preserving the predictive effect on the trailer motion.

This algorithm enables accurate path-tracking without requiring full MPC optimization, offering a compromise between computational efficiency and prediction-based robustness.

The algorithm is presented in Algorithm~\ref{alg:path_tracking}. It relies on three components that can easily be replaced by alternatives: the lateral control law $f_{lat}$, the minimization strategy used to compute $\delta_k$, and the model update function $ModelUpdate$.

In this paper, $f_{lat}$ is the controller introduced in \cite{lombard2026path}, while $ModelUpdate$ is based on the extended kinematic bicycle model to simulate the evolution of the tractor-trailer system over time. A key advantage of the controller in \cite{lombard2026path} is that it provides better accuracy and disturbance tolerance than most state-of-the-art methods, while allowing the control point at which the command is applied to be chosen (i.e., either the front or the rear of the trailer can be made to follow the trajectory, improving the maneuverability of the TTS).

\begin{algorithm}[!htbp]
\caption{Tractor--Trailer Path-Tracking Iterative Control}
\label{alg:path_tracking}
\begin{algorithmic}[1]

\REQUIRE Initial tractor and trailer state $X_0$, number of iterations $N$
\ENSURE Steering command $\bar{\delta}$

\STATE Initialize sum of steering commands: $S \gets 0$
\STATE Initialize state: $X \gets X_0$

\FOR{$k = 1$ to $N$}
    \STATE Compute desired articulation using trailer lateral control:
    \[
        \phi_{\mathrm{des}} \gets f_{\text{lat}}(X)
    \]
    \STATE Find steering angle minimizing the articulation error:
    \[
        \delta_k = \arg\min_{\delta} \left| \phi_{\mathrm{des}} - \phi_k(\delta) \right|
    \]
    \STATE Update tractor--trailer state using $\delta_k$:
    \[
        X \gets \text{ModelUpdate}(X, \delta_k)
    \]
    \STATE Accumulate steering commands: $S \gets S + \delta_k$
\ENDFOR

\STATE Compute final averaged steering angle:
\[
    \bar{\delta} \gets \frac{S}{N}
\]

\RETURN $\bar{\delta}$

\end{algorithmic}
\end{algorithm}

\subsection{Correction maneuvers}

As shown by the articulation dynamics \eqref{eq:articulation_dynamics}, the reachability of a target articulation angle depends on the driving direction.

In forward maneuvers ($v>0$), for any desired articulation angle $\phi^\star$, an equilibrium of \eqref{eq:articulation_dynamics} exists and is obtained by choosing the tractor curvature as
\begin{equation}
\kappa^\star = \frac{1}{L_2}\sin\phi^\star.
\label{eq:forward_equilibrium}
\end{equation}
Provided that the corresponding steering command satisfies the actuator constraints, i.e., $|\kappa^\star|\leq \kappa_{\max}$, the articulation angle $\phi^\star$ is kinematically reachable. Moreover, since $\dot{\phi}$ is monotonic with respect to $\kappa$ when $v>0$, the steering angle can be adjusted such that $\dot{\phi}$ has the appropriate sign to drive $\phi$ toward $\phi^\star$. This property ensures that, \textbf{in forward motion, the target tractor--trailer articulation angle can always be reached within the limits imposed by steering saturation}.

In backward maneuvers ($v<0$), the sign of the articulation dynamics in \eqref{eq:articulation_dynamics} is reversed, which fundamentally alters the reachability properties. For instance, to reduce a positive articulation angle $\phi>0$ toward zero, it is necessary to satisfy
\begin{equation}
\dot{\phi} < 0 \;\;\Longleftrightarrow\;\; \kappa > \frac{1}{L_2}\sin\phi.
\label{eq:backward_condition}
\end{equation}
If the maximum achievable curvature $\kappa_{\max}$ satisfies
\begin{equation}
\kappa_{\max} \leq \frac{1}{L_2}\sin\phi,
\end{equation}
then no admissible steering command can decrease the articulation angle. In this case, the articulation angle either remains constant or increases, leading to a jackknife-prone configuration. Consequently, \textbf{in backward motion, it is not always possible to reach or even reduce the tractor--trailer articulation angle toward a desired target due to the combined effect of the reversed kinematics and steering constraints}.

\begin{figure}
    \centering
    \includegraphics[width=0.75\linewidth]{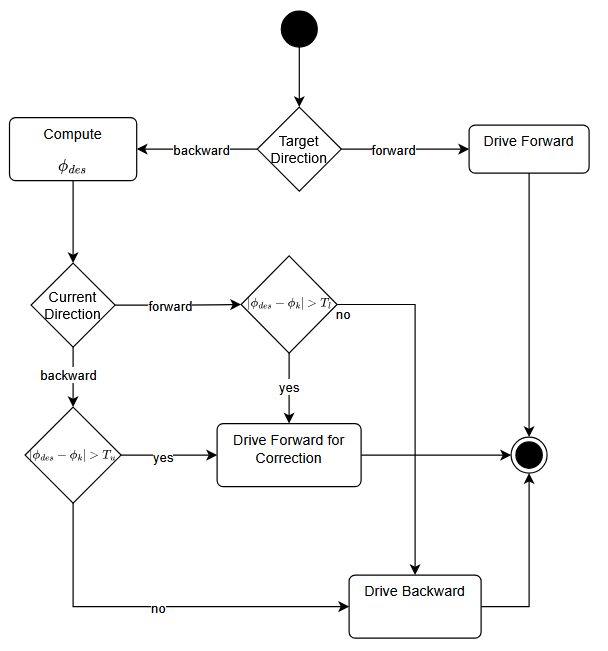}
    \caption{Statechart of the selection of a driving mode according to $\phi_{des}$}
    \label{fig:driver-statechart}
\end{figure}

To overcome this limitation, a correction maneuver in forward motion is introduced whenever the desired articulation angle cannot be reached in backward motion. During this maneuver, the system exploits the favorable kinematic properties of forward driving to bring the articulation angle to a feasible target value. Once the desired articulation angle is reached, the backward maneuver can be safely resumed. This process of selecting the appropriate driving mode is described in Fig. \ref{fig:driver-statechart}.

The main idea of the process described in Fig.~\ref{fig:driver-statechart} is to determine whether backward motion allows the articulation angle to evolve toward the desired target
$\phi_{des}$. Since $\phi_{des}$ is an instantaneous geometric target, the controller evaluates the deviation $|\phi_{des} - \phi|$ to assess articulation recoverability under steering saturation. When this deviation exceeds a threshold $T$, no admissible steering command can orient the articulation dynamics toward $\phi_{des}$ in backward motion. A forward correction maneuver is then triggered to exploit the favorable reachability properties of forward driving. Once $|\phi_{des} - \phi| \leq T$, backward motion can be safely resumed.

In addition, to prevent oscillations caused by the proximity between $|\phi_{des} - \phi|$ and $T$ (which would cause the driver to correct forward, then go backward as soon as $|\phi_{des} - \phi| \leq T$), the threshold is separated into $T_u$ and $T_l$, with $T_u$ being the threshold used to begin the correction maneuver, and $T_l$ the threshold to stop the correction maneuver, where $T_u > T_l$. These two thresholds set the trade-off between articulation accuracy and the number and duration of correction maneuvers. A high value of $T_u$ makes the controller more tolerant to deviations from the target articulation angle before a correction is triggered, whereas a low value of $T_u$ triggers corrections as soon as small deviations occur. Conversely, a high value of $T_l$ interrupts the correction early, whereas a low value of $T_l$ requires the trailer to be accurately aligned with the target articulation angle before backward motion is resumed, at the cost of longer correction maneuvers.

\section{Experimentation}
\label{sec:experimental_setup}

\subsection{Simulation platform and scenario}

The algorithm was first validated in a lightweight simulation environment modeling a tractor towing a single-axle trailer, developed within the xHUB project and available at \url{https://fperronn.github.io/xhub/dhl/}. Several scenarios were tested, including sharp turns, high-curvature paths, and abrupt perturbations.

\begin{figure}
    \centering
    \includegraphics[width=0.6\linewidth]{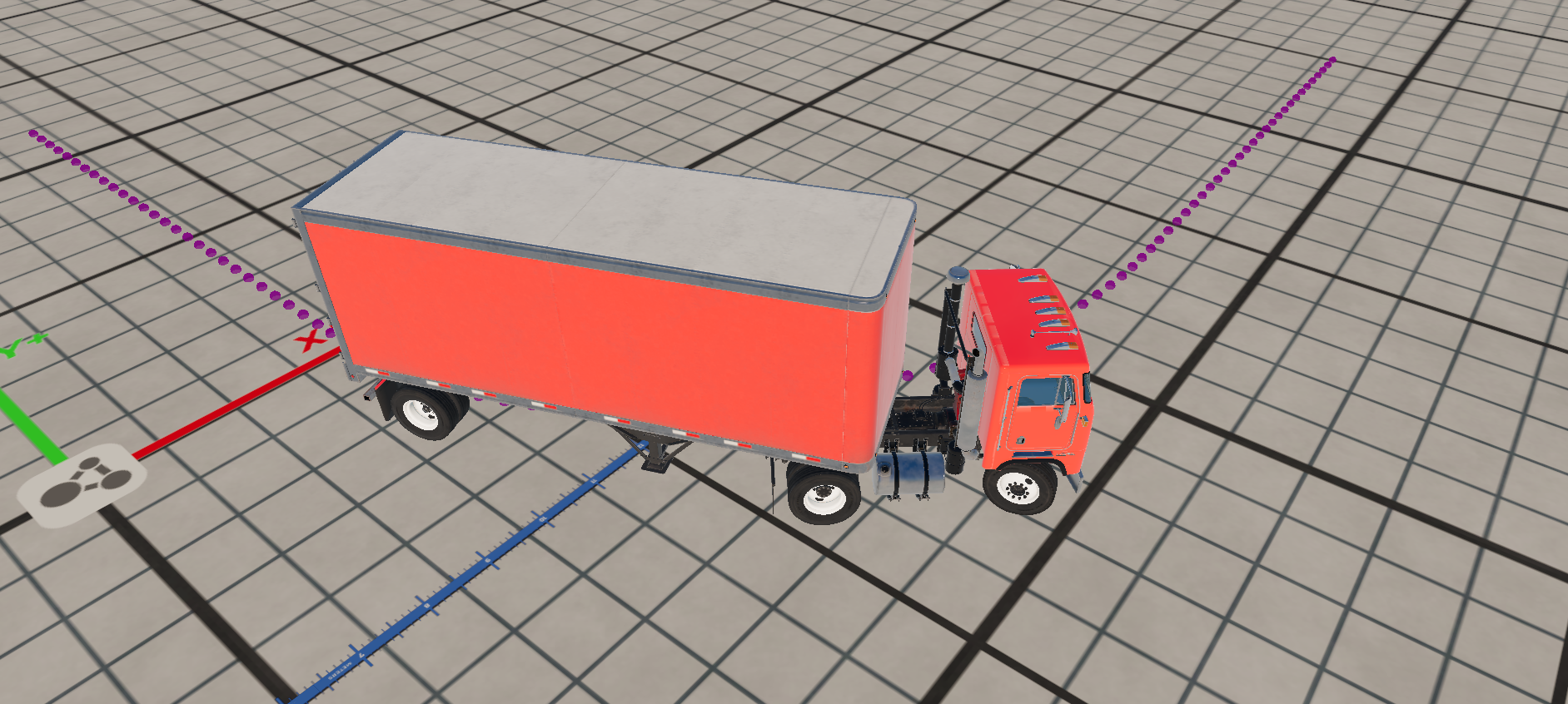}
    \caption{Screenshot of the BeamNG.tech simulator of the TTS following a trajectory for a docking maneuver}
    \label{fig:beamng}
\end{figure}

Since this environment relies on the extended bicycle model itself, further experiments were conducted with the BeamNG.tech simulator \cite{maul2021beamng} (Fig.~\ref{fig:beamng}), which provides a higher-fidelity physical simulation that includes lateral tire forces and thus tests the controller beyond its own modeling assumptions.

Three scenarios were chosen to validate the behavior of the controller: a straight line, a circle, and a reverse parking (docking) maneuver, each following a predefined trajectory generated with a quintic Hermite spline. The speed during the maneuvers was limited to approximately $1$~m/s, which is a typical speed for docking maneuvers, and the selected control point was the center of the trailer axle.

Using the BeamNG.tech simulator, experiments were also conducted with trailers of various lengths, geometries (in particular different numbers of axles and tires), and weights.

\subsection{Experimental results}

The geometric parameters of the tractor-trailer system are reported in Table~\ref{tab:sim_params}, where the hitch offset $h_1$ denotes the signed distance between the hitch point and the middle of the rear axle of the tractor.
\begin{table}[t]
    \centering
    \begin{tabular}{l|c}
         \textbf{Parameter} & \textbf{Value}  \\
         Tractor wheelbase & 2.93~m \\
         Trailer wheelbase & 6.30~m \\
         Hitch offset ($h_1$ in Fig.~\ref{fig:extended_bicycle}) & 0.14~m \\
         Steering angle rate limit & 40\textdegree/s \\
    \end{tabular}
    \caption{Simulation parameters}
    \label{tab:sim_params}
\end{table}

Two metrics were measured over time: the lateral error at the control point (m) and the heading error of the trailer at the control point (rad).

\begin{figure}
\centering
\includegraphics[width=0.6\linewidth]{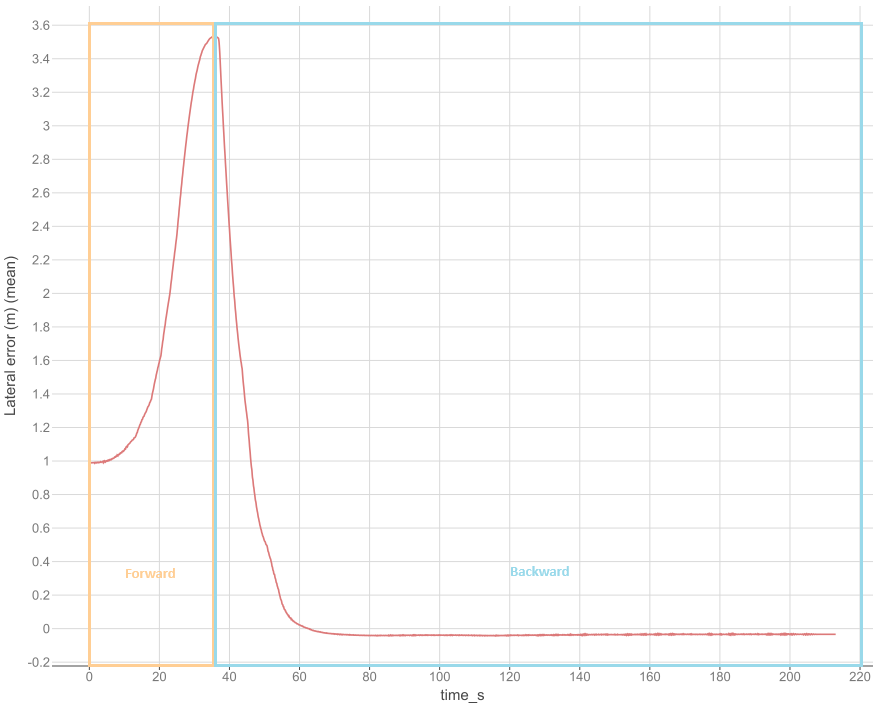}
\caption{Lateral error (m) for the straight-line trajectory.}
\label{fig:line_lat}
\end{figure}

\begin{figure}
\centering
\includegraphics[width=0.6\linewidth]{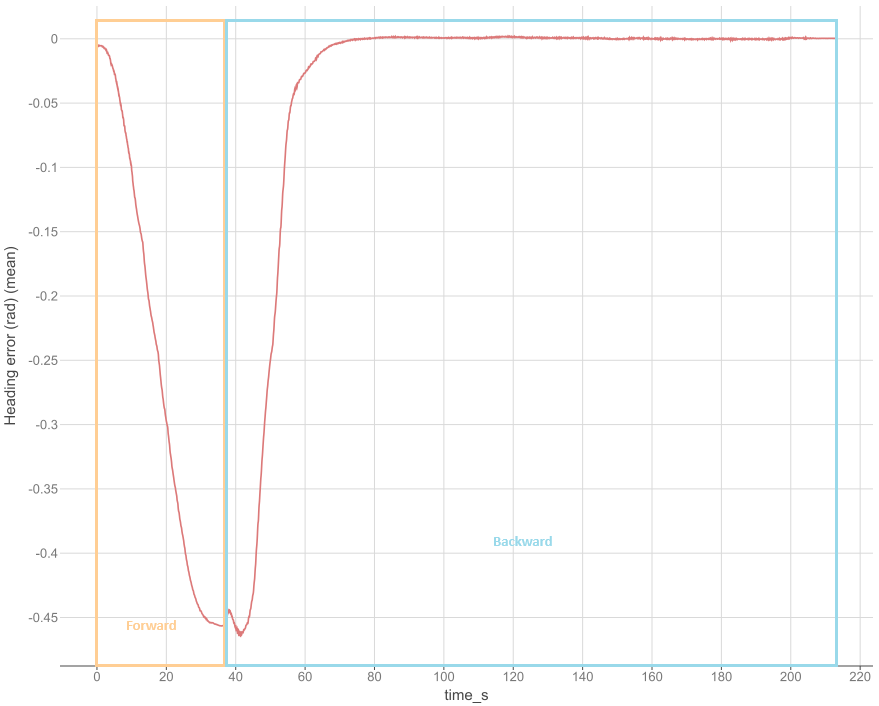}
\caption{Heading error (rad) for the straight-line trajectory.}
\label{fig:line_heading}
\end{figure}

In Fig.~\ref{fig:line_lat}, the vehicle is spawned next to the trajectory and rejoins it during the first 60~s. This phase is divided in two stages: from 0~s to 30~s the vehicle drives forward to obtain a better angle for the backward maneuver, then from 30~s to 60~s it approaches the trajectory while reversing. After 60~s, the vehicle follows the line in steady state with a stable lateral error of about $0.03$~m and a heading error stable below $10^{-4}$~rad (Fig.~\ref{fig:line_heading}).

\begin{figure}
\centering
\includegraphics[width=0.6\linewidth]{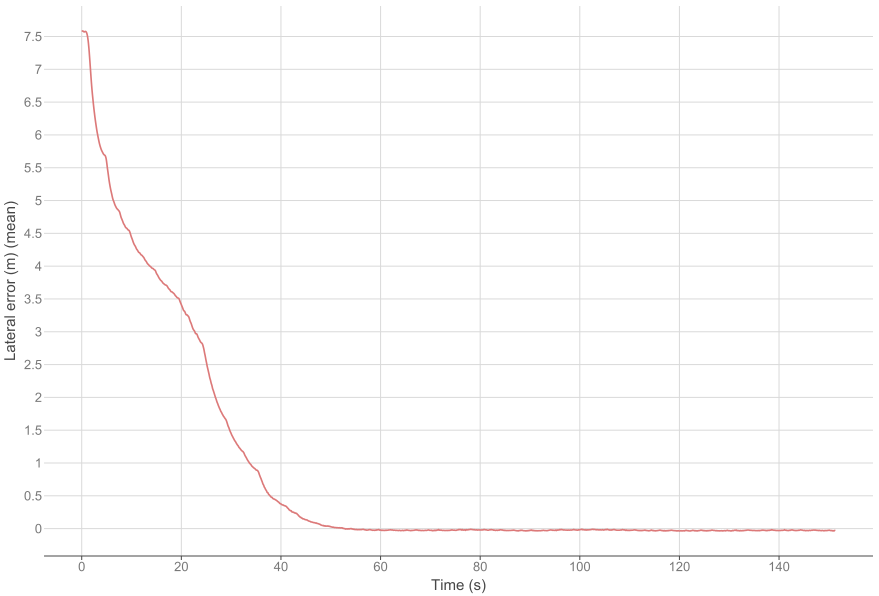}
\caption{Lateral error (m) for the circular trajectory.}
\label{fig:circle_lat}
\end{figure}

\begin{figure}
\centering
\includegraphics[width=0.6\linewidth]{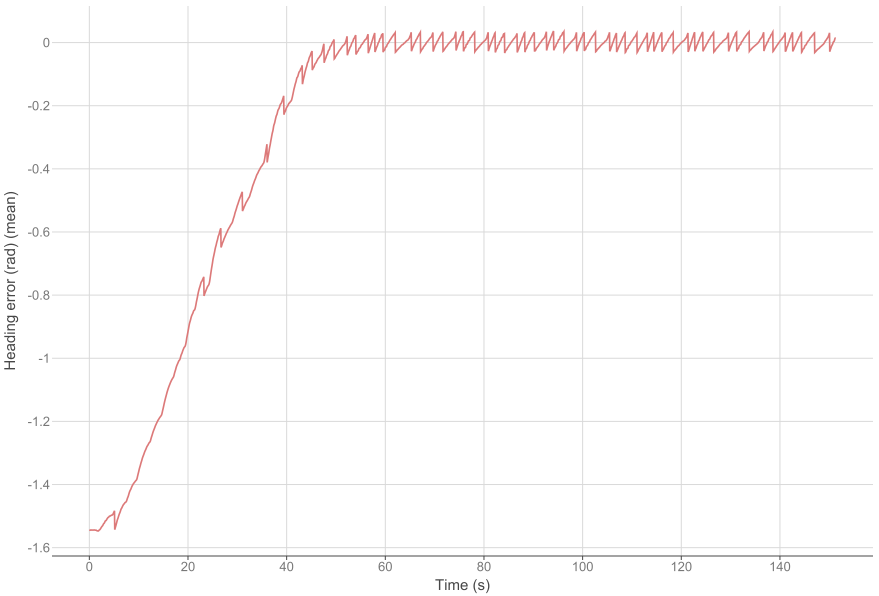}
\caption{Heading error (rad) for the circular trajectory.}
\label{fig:circle_heading}
\end{figure}

In Fig.~\ref{fig:circle_lat}, the vehicle drives directly toward the circular trajectory. After about 50~s it reaches the trajectory and follows it with a steady-state lateral error of about $0.02$~m and a heading error below $0.03$~rad (Fig.~\ref{fig:circle_heading}).

\begin{figure}
\centering
\includegraphics[width=0.6\linewidth]{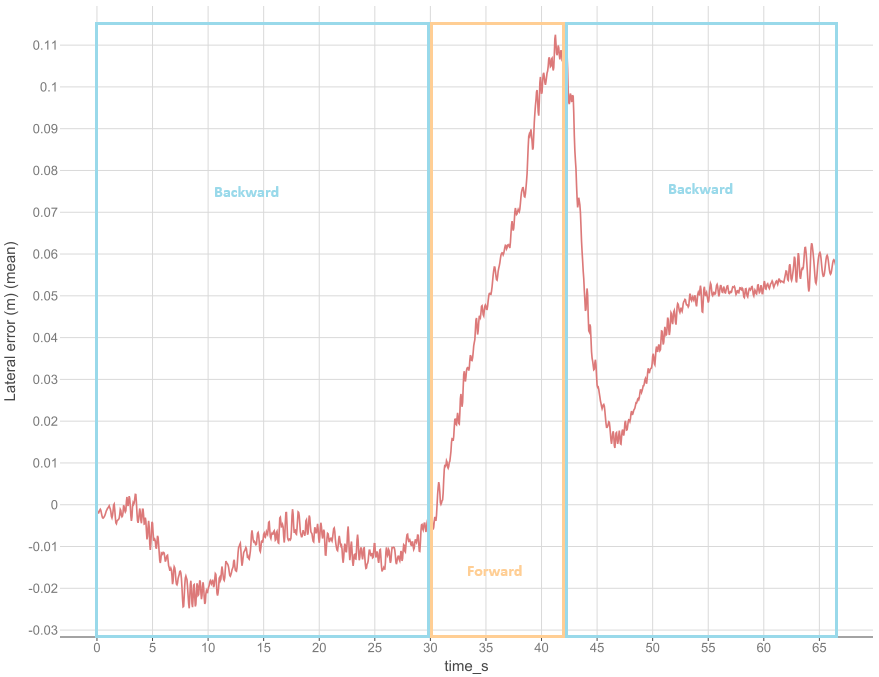}
\caption{Lateral error (m) for the docking trajectory. The shaded bands indicate the forward and backward driving phases.}
\label{fig:docking_lat}
\end{figure}

\begin{figure}
\centering
\includegraphics[width=0.6\linewidth]{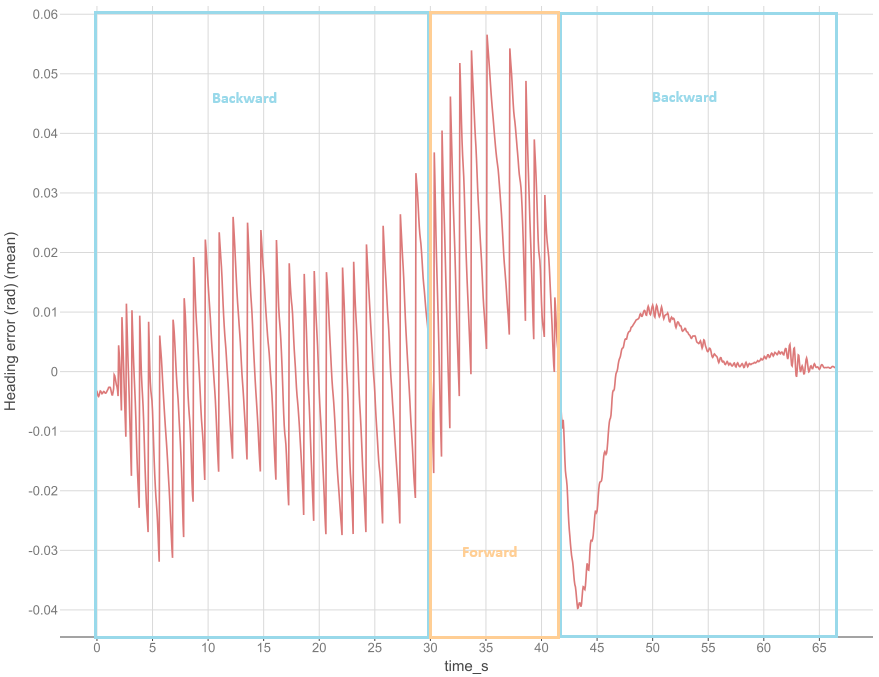}
\caption{Heading error (rad) for the docking trajectory.}
\label{fig:docking_heading}
\end{figure}

The docking maneuver is the most challenging scenario because of its varying curvature (Fig.~\ref{fig:docking_lat}). The vehicle is spawned on the trajectory, but the sharp curve triggers a forward correction maneuver around 30~s, which temporarily increases the lateral error as the vehicle drives forward to recover a feasible alignment. At the end of the maneuver, the vehicle is aligned with the trajectory (Fig.~\ref{fig:docking_heading}, heading error below $0.01$~rad) while being about $0.05$~m from it.

We note that the low-level longitudinal control (regulating pedal pressure to reach a target acceleration) was difficult to tune: abrupt velocity changes perturbed the trajectory and degraded performance, and better results can be expected with an improved longitudinal controller \cite{liu2023experimental}.

Across all scenarios, the proposed controller achieved low lateral error and smooth articulation dynamics. Since the vehicles, scenarios, and control parameters used in the literature differ from ours, a direct comparison is not possible; the mean errors reported in other works are therefore given only as an indication of the expected order of magnitude:
\begin{itemize}
    \item In \cite{manav2021adaptive}, the maximum lateral error is in $[0.04, 0.08]$~m in straight-line motion with a steady-state error close to $0$~m, while the steady-state lateral error reaches $0.12$~m in circular motion.
    \item In \cite{bin2023sliding}, the mean lateral error is around $0.33$~m (max around $0.45$~m), with a heading error of about $0.1$~rad.
\end{itemize}

The errors obtained with the proposed controller are therefore of the same order of magnitude as, or lower than, those reported for state-of-the-art controllers. Moreover, it was successfully applied to various TTS geometries (not shown here for space reasons) without any issue; its only parameters are the tractor and trailer wheelbases and the hitch offset $h_1$, and no specific calibration or training was required to adapt to the different geometries.

\subsection{Real-world experiments}
\label{subsec:real_experiments}

\begin{figure}
    \centering
    \includegraphics[width=0.6\linewidth]{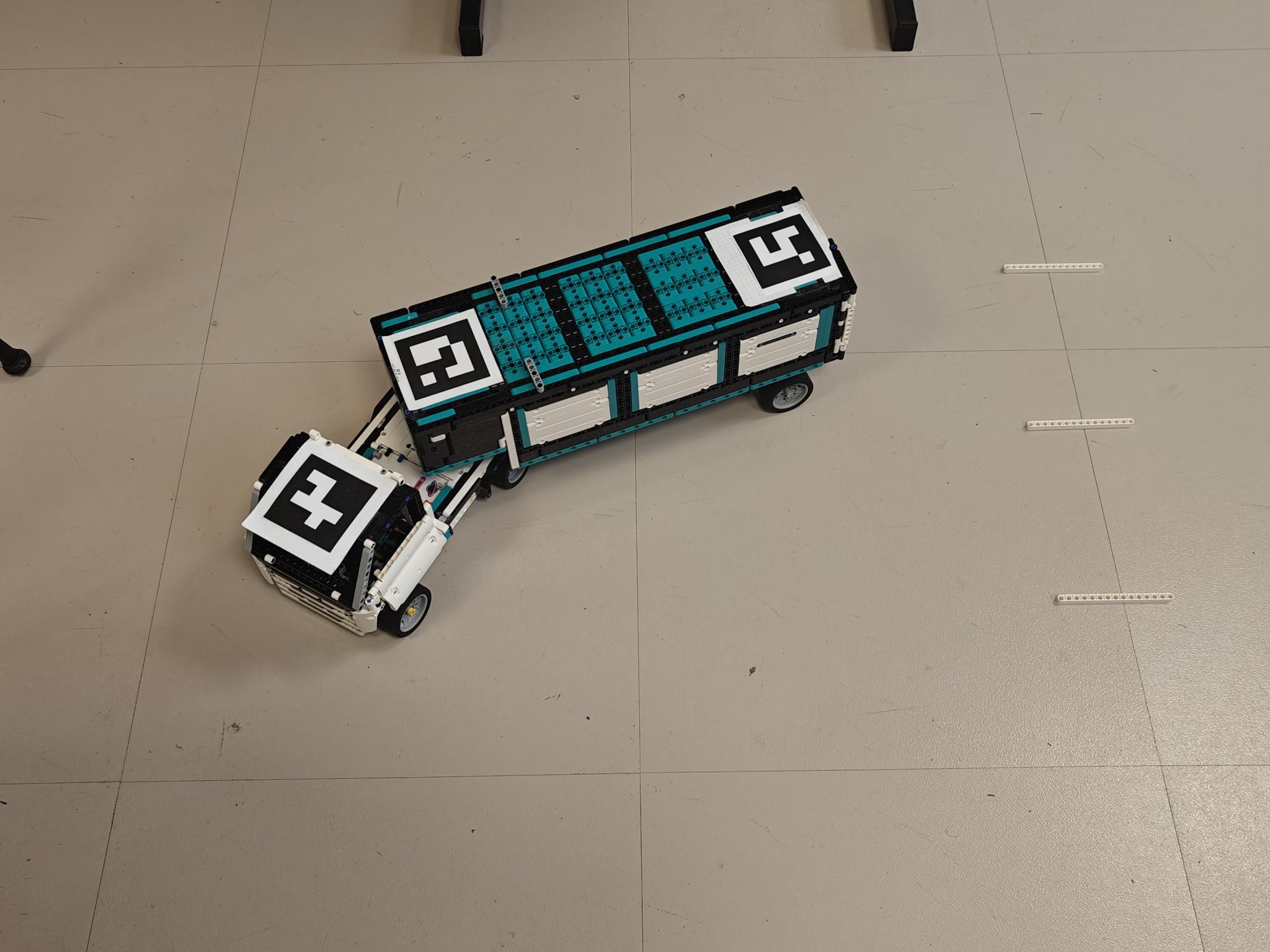}
    \caption{Small-scale tractor--trailer robot built with LEGO Mindstorms Robot Inventor. The markers placed on the tractor and the trailer are tracked by an external ZED~2 camera.}
    \label{fig:lego}
\end{figure}

To assess the feasibility of the approach on a physical system, subject to sensor noise, actuator imperfections, and limited computational resources, the controller was also deployed on a small-scale tractor--trailer robot built with LEGO Mindstorms Robot Inventor (Fig.~\ref{fig:lego}). The poses of the tractor and of the trailer are estimated by an external ZED~2 stereo camera, which detects ArUco markers placed on both bodies, and are transmitted to the robot via Bluetooth. The whole controller (trailer-level lateral control, predictive steering computation, and driving-mode selection) runs on the LEGO hub itself, a microcontroller-based brick (STM32F413, ARM Cortex-M4 at 100~MHz). The robot has a tractor wheelbase of $0.202$~m, a trailer wheelbase of $0.365$~m, a hitch offset $h_1$ of $0.024$~m, and a maximum steering angle of $37.0$\textdegree.

The scenario consists of a reverse docking maneuver into two parallel bays separated by $0.20$~m. Each cycle is composed of a backward maneuver of about $2$~m from the starting position into one of the bays, followed by a forward maneuver of the same length back to the starting position. The reference path has a minimum radius of curvature of $0.48$~m. Ten consecutive cycles were performed, alternating between the two bays (about 15~min of continuous operation). Estimated from the measured trailer positions, the mean speed was about $0.04$~m/s in backward motion and $0.05$~m/s in forward motion, and never exceeded $0.07$~m/s.

The results are summarized in Table~\ref{tab:lego_results}. The lateral error at the control point, as measured by the camera-based localization, remains below $13$~mm in backward motion and below $7$~mm in forward motion, while the articulation angle reaches up to $0.58$~rad during reversing. At the end of each docking maneuver, the trailer is positioned within $5.1$~mm of the reference path with a heading error below $0.006$~rad, and these final positions are consistent across the five repetitions performed for each bay. These results were obtained despite a pose detection latency of about 200~ms, which shows the robustness of the controller to such delays.

\begin{figure}
    \centering
    \includegraphics[width=0.6\linewidth]{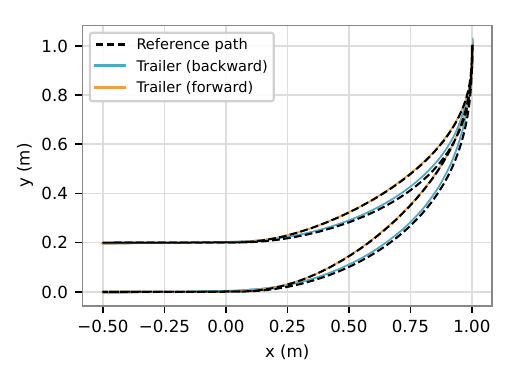}
    \caption{Reference path and trajectories of the trailer control point of the LEGO TTS over the ten docking cycles.}
    \label{fig:lego_trajectories}
\end{figure}

\begin{figure}
    \centering
    \includegraphics[width=0.6\linewidth]{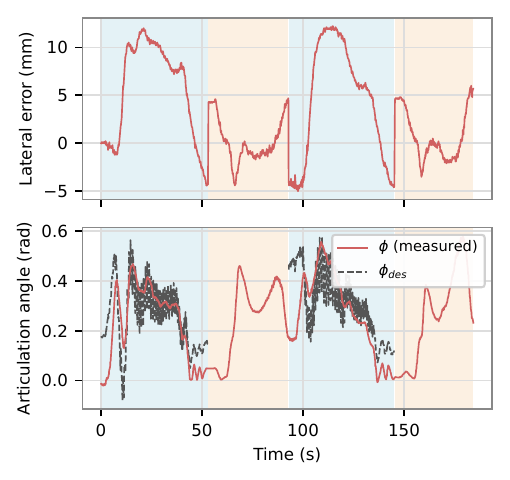}
    \caption{Lateral error and articulation angle of the LEGO TTS during the first two docking cycles. The shaded bands indicate the backward (blue) and forward (orange) driving phases; the desired articulation angle $\phi_{des}$ is shown in backward motion.}
    \label{fig:lego_errors}
\end{figure}

Fig.~\ref{fig:lego_trajectories} shows the trajectories of the trailer control point over the ten cycles: they are almost superimposed on the reference path, the tracking errors being of a few millimeters at the scale of the maneuver. Fig.~\ref{fig:lego_errors} details the first two cycles, one for each bay. In backward motion, the lateral error reaches about $12$~mm when the trailer follows the curved part of the path, and decreases as the path becomes straight when approaching the bay. The desired articulation angle $\phi_{des}$ computed by the trailer-level lateral control law is affected by the noise of the camera-based localization, whereas the measured articulation angle remains smooth.

\begin{table}[t]
    \centering
    \begin{tabular}{l|c|c}
         & \textbf{Backward} & \textbf{Forward} \\
         \hline
         RMS lateral error (mm) & 7.5 & 2.5 \\
         Max. lateral error (mm) & 13.2 & 6.1 \\
         RMS heading error (rad) & 0.021 & 0.022 \\
         Max. articulation angle (rad) & 0.58 & 0.63 \\
    \end{tabular}
    \caption{Tracking errors of the LEGO TTS over 10 docking cycles}
    \label{tab:lego_results}
\end{table}

Regarding computational cost, the duration of each control iteration was measured on the PC as a complete round trip: transmission of the pose detected by the camera to the robot, decoding of the Bluetooth message, acquisition of the robot sensor data, computation of the command, transmission of the command to the motors, and reception by the PC of the acknowledgment sent by the robot. This round-trip time, which is therefore an upper bound of the time required to compute the command on the LEGO hub, has a median of $62$~ms (about $16$~Hz), and $99\%$ of the iterations are completed within $181$~ms. This shows that the proposed controller is compatible with low-cost embedded platforms, without requiring a dedicated computing unit.

\section{Conclusion}
\label{sec:conclusion}

This paper introduces a novel tractor-trailer path-tracking algorithm based on a two-stage control strategy combining trailer-level lateral guidance and predictive steering adjustment, complemented by a forward correction maneuver that aims to restore the recoverability of the articulation angle during reversing. The method is model-aware and computationally efficient, as its whole control loop runs at about 16~Hz on a low-cost microcontroller. Being geometry-based, it adapts to various TTS geometries without any specific calibration. The simulations performed with the BeamNG.tech simulator show tracking errors of the same order of magnitude as, or lower than, those reported for state-of-the-art controllers, and the experiments on a small-scale robot confirm its feasibility under real sensing and actuation conditions, with lateral errors below 13~mm during repeated reverse docking maneuvers despite a pose detection latency of about 200~ms.

Future work will address several directions. First, the controller will be extended with a longitudinal control layer, in order to manage the vehicle speed jointly with the steering command. Second, the robustness of the controller to disturbances, measurement noise, latencies, and model parameter errors will be studied more thoroughly: although the results obtained in this work are encouraging, they do not constitute a formal demonstration of robustness. Third, more detailed comparisons with other state-of-the-art solutions will be performed. Finally, experiments on a full-scale truck, which are currently in progress, will evaluate the performance of the controller in the presence of the sensor and actuator inaccuracies and latencies of real vehicles.

\section*{Acknowledgements}
This work was supported by ANR/xHUB. The authors thank BeamNG GmbH for providing an academic license of BeamNG.tech.

\section*{CRediT authorship contribution statement}
\textbf{Alexandre Lombard:} Conceptualization, Methodology, Validation, Formal analysis, Investigation, Visualization, Writing -- original draft, Writing -- review \& editing, Supervision.
\textbf{Florent Perronnet:} Conceptualization, Software, Investigation, Writing -- review \& editing.
\textbf{Th\'eo Rigal:} Software, Investigation, Data curation, Writing -- review \& editing.
\textbf{Nicolas Gaud:} Resources, Supervision, Writing -- review \& editing.
\textbf{Abdeljalil Abbas-Turki:} Resources, Supervision, Project administration, Funding acquisition, Writing -- review \& editing.

\section*{Declaration of competing interest}
The authors declare that they have no known competing financial interests or personal relationships that could have appeared to influence the work reported in this paper.

\section*{Data availability}
Data will be made available on request.

\section*{Declaration of generative AI and AI-assisted technologies in the manuscript preparation process}
During the preparation of this work the authors used Claude (Anthropic) in order to assist with the revision of the manuscript text and the generation of figures from the experimental data. After using this tool, the authors reviewed and edited the content as needed and take full responsibility for the content of the published article.

\bibliographystyle{elsarticle-num}
\bibliography{biblio}

\end{document}